\documentclass{article}
\usepackage{stywhispers,amsmath,epsfig}
\usepackage{amssymb}
\usepackage{amsmath}
\usepackage{amsthm}
\usepackage{bm}
\usepackage{multirow}
\usepackage{nicematrix}
\usepackage{xcolor, colortbl}
\usepackage{ulem}
\usepackage{highlight}
\usepackage[colorlinks=true]{hyperref}
\newcommand{\ApplyGradient}[1]{%
        \ifdim #1 pt > \MidNumber pt
            \pgfmathsetmacro{\PercentColor}{max(min(100.0*(#1 - \MidNumber)/(\MaxNumber-\MidNumber),100.0),0.00)} %
            \hspace{-0.33em}\colorbox{green!\PercentColor!yellow}{#1}
        \else
            \pgfmathsetmacro{\PercentColor}{max(min(100.0*(\MidNumber - #1)/(\MidNumber-\MinNumber),100.0),0.00)} %
            \hspace{-0.33em}\colorbox{red!\PercentColor!yellow}{#1}
        \fi
}
\newcolumntype{R}{>{\collectcell\ApplyGradient}c<{\endcollectcell}}

\title{SimPINNs: Simulation-Driven Physics-Informed Neural Networks for Enhanced Performance in Nonlinear Inverse Problems}
\name{Sidney Besnard$^{1,2}$  , Frederic Jurie$^1$, Jalal Fadili$^1$ }
\address{$^1$Univ. Caen Normandie, ENSICAEN, CNRS \hspace{1cm}   $^2$Safran Data Systems}

\begin{document}
%\ninept
%
\maketitle

\begin{abstract}
This paper introduces a novel approach to solve inverse problems by leveraging deep learning techniques. The objective is to infer unknown parameters that govern a physical system based on observed data. We focus on scenarios where the underlying forward model demonstrates pronounced nonlinear behaviour, and where the dimensionality of the unknown parameter space is substantially smaller than that of the observations. Our proposed method builds upon physics-informed neural networks (PINNs) trained with a hybrid loss function that combines observed data with simulated data generated by a known (approximate) physical model. Experimental results on an orbit restitution problem demonstrate that our approach surpasses the performance of standard PINNs, providing improved accuracy and robustness.
\end{abstract}
\begin{keywords}
Inverse problems, Neural Networks, Physics-Informed, Simulation  
\end{keywords}
\section{Introduction}
\label{sec:intro}

Inverse problems play a crucial role in science by allowing to unravel the hidden properties and processes behind observed data. They allow scientists to infer and understand phenomena that are otherwise difficult or impossible to observe or measure directly. These problems involve determining the parameters of a system from some available measurements. Inverse problems have far-reaching applications spanning a wide spectrum ranging from medical imaging to non-destructive control or astronomical imaging, as we will see.

Machine and deep learning have recently emerged as powerful alternatives to variational models for solving inverse problems. These methods include supervised and unsupervised methods, such as Deep Inverse Prior (DIP) \cite{Ulyanov:2020}, Unrolling \cite{Peng:2019,Gregor:2010}, Plug-and-play (PnP) \cite{Venkatakrishnan:2013,Wu:2020}, and generative models \cite{Bora2017}. For a review see \cite{arridge_maass_öktem_schönlieb_2019}. Unrolling and PnP rely on neural networks to learn the regularization from the data.

However, these approaches only make sense if the output parameter space can be equipped with a suitable notion of regularity. This is certainly the case if the input parameters are in the form of a structured signal, but is not always the case as in our setting (think of inferring a few parameters that are not structured on a grid). A naive technique would then be to train a neural network using a dataset consisting of input-output pairs, where the input is the observed data and the output is the sought-after vector of parameters \cite{Metzler:2020, Metzler:2018, Rivenson:2017}. Clearly, the neural network learns to invert the forward model (i.e. the mapping between the observed data and parameters), with the hope that it would predict the unknown parameters for new observations. This approach leverages the ability of neural networks to capture intricate patterns and non-linear relationships in the data. Unfortunately, this type of technique is only applicable when a large set of training pairs is available, which is barely the case in most practical situations. Moreover, such approaches are completely agnostic to the forward model which would produce unrealistic solutions and may not generalize well.

Physics-informed neural networks (PINNs) were primarily proposed to solve partial differential equations (PDE) \cite{RAISSI2019686, Raissi2017PhysicsID, Uddin:2023, Cai:2021}. Their core idea is to supplement the neural network training with information stemming from the measurement formation model, e.g. the PDE model. In turn, this allows to restrict the space of solutions by enforcing the output of the trained neural network to comply with the physical model as described by the PDE. In turn, these methods are expected to be trained with a smaller dataset.

An aspect to consider regarding PINNs is their reliance solely on the reconstruction error, which represents constraints imposed by the partial differential equation (PDE). However, in numerous cases, achieving a low reconstruction error does not guarantee an accurate prediction of the parameters (i.e., a low parameter error). Thus, it is essential to emphasize the network's requirement for regularization of the solution space during training.

% This led to our method, which regularises the solution space directly from labelled data. Drawing inspiration from self-supervised physics-informed neural networks \cite{subramanian2022adaptive}, this method adds a supervised term in the loss function to guide the network and help to regularise the problem.

% The use of physics-informed neural networks (PINNs)\cite{RAISSI2019686, Raissi2017PhysicsID, Uddin:2023, Hao:2023} for solving inverse problems has gained significant attention in recent years. PINNs combine the strengths of physics-based modelling and deep learning to address inverse problems in a more robust and interpretable manner. In PINNs, the neural network architecture is augmented with the governing equations or physical constraints that govern the underlying system. By incorporating these known physics principles into the training process, PINNs can effectively learn the relationship between the observed data and the unknown parameters while satisfying the fundamental laws of the system. This physics-informed training provides a regularization mechanism, reducing the reliance on large amounts of training data.. The combination of physics-based constraints and deep learning allows for more accurate and reliable solutions to inverse problems while leveraging the domain knowledge encoded in the underlying physics. Such approaches are generally based on observational data only, and therefore have the advantage of not requiring training data pairs. 

\paragraph*{Contributions.}
In this paper, we propose a novel hybrid approach that combines both physics-informed and data-driven methods by using simulated data to regularise the solution space. Given the difficulty in obtaining real training pairs (observations, parameters) for many real-world problems, simulations offer a convenient means to provide the missing complementary information. Our aim is to demonstrate the effectiveness of neural networks in dealing with non-linear problems, while presenting a method that implicitly infers the forward operator within the neural network parameters by minimising the reconstruction error. 

% Through our investigation, we will show that leveraging the information provided by physics is sufficient to achieve model convergence and generalization without requiring labelled data (PINNS approach). In addition, we introduce a hybrid version that balances the tradeoff between data fidelity and physics fidelity by minimizing both the reconstruction error and the parameters error on simulated data. We show that this simulation-driven physics-informed neural network improves performance even further, especially when observations are limited. This observation is true even when the physical model used in the simulation is only known approximately. 

\begin{figure}[tb]
\begin{center}
\includegraphics[width=.5\textwidth]{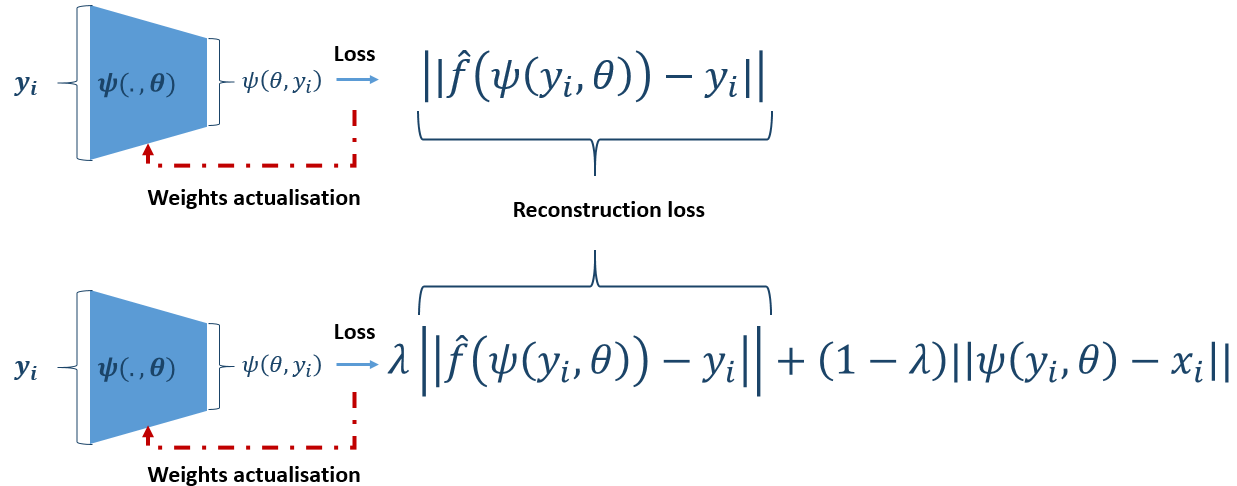}
\end{center}
% \caption{Proposed models architecture.}
\caption{Illustrations of the architectures used in PINNs and SimPINNs for inverse problems. The top part depicts the PINNs approach, where a neural network is trained to learn the inverse function of $f$, enabling the reconstruction of accurate $x$ values from observed $y$ values solely based on the observations and the underlying physics (unsupervised learning). In the bottom part, the SimPINNs (supervised) approach is shown, which utilizes the physics-based simulations to complement the training set with 'annotated' simulated data and regularize the solution.  \label{fig:proposed_models}}
\end{figure}

\section{Proposed method\label{sec:proposed_method}}
%\subsection{Problem formalization and notations}

Let $\mathcal{X} \subset \mathbb{R}^n$ be  the space of parameters of the (physical) model, and $\mathcal{Y} \subset \mathbb{R}^m$ be the space of observations. An inverse problem consists in reliably recovering the parameters $x \in \mathcal{X}$ from noisy indirect observations
\begin{align}\label{eq:forward}
y = f(x) + \epsilon ,
\end{align}
where $f:\mathcal{X}\to \mathcal{Y}$ is the forward operator, and $\epsilon$ stands for some additive noise that captures the measurement and possibly the modeling error. Throughout, we assume that $f$ is smooth enough (at least continuously differentiable).

In the sequel, for a neural network with parameters (weights and biases) $\theta \in \Theta$, $\psi: (y,\theta) \in \mathcal{Y} \times \Theta \mapsto \psi(y,\theta)$ denotes its output.

Finally, we also assume that we have an approximate explicit model of $f$, sometimes referred to as a digital twin. This model, $\hat{f}$, is obtained by modelling the physical phenomena involved in the observations. It will be used later to generate simulated data.

\subsection{PINNs for non-linear inverse problems}

% As one can see, the methods introduced in the previous section are focused on  the regularization of the solution space, rather than the  data fidelity term  $||f(x)-y||_2^2$ that is convex in the case of a linear forward operator. 

% Under the condition that the forward operator is non-linear, equation \ref{eq:grad_loss} will push the solution onto a local minimal of the data fidelity term which is not satisfying for us, i.e., the reconstruction error will stay high.

The key idea of PINNs is to incorporate the physical model into the cost function during the training process. For a neural network $\psi$ and training samples $\{y_i: i=1,\ldots,n\}$ with $n$ samples, this amounts to solving the following minimization problem with the empirical loss:
\begin{equation}\label{eq:trainpinnunsup}
\min_{\theta \in \Theta} \frac{1}{n}\sum_{i=1}^n \lVert y_i - \hat{f}(\psi(y_i,\theta)) \rVert^2
\end{equation}
This loss function leverages the information provided by the physical forward model directly into the training loss. It is also a non-supervised method that relies solely on observations, without any knowledge of the parameter vector $x_i$ corresponding to each $y_i$. Unfortunately, as was observed previously in the literature (e.g. in \cite{Chen21}), when $f$ is not injective, there are infinitely many solutions $\psi(\cdot,\theta)$ which attain zero training error. This is because the forward model $f$ may map multiple input vectors to the same output vector. For example, in the linear case, the action of $f$ is invariant along its null space. This suggests that training a reconstruction network as \eqref{eq:trainpinnunsup} only from the observed data, without any additional assumptions or constraints, is not viable. Possible workarounds include explicitly constraining the output of the reconstruction network through regularization (and we are back to the variational world), or introducing invariances such as in \cite{Chen21}. In the forthcoming section, we will describe an alternative based on exploiting the forward model to simulate input-output pairs. This approach will help to regularize the training process and make it more robust to the non-injectivity of $f$.

\subsection{SimPINNs: Simulation aided PINNs}
In many areas of science, obtaining pairs of input (parameters)-output (observations) training data, can be a significant challenge. This can be due to various reasons, including that data are difficult or expensive to acquire. This is for instance the case in large instruments in physics. Furthermore, even if such pairs of data can be acquired, they are available only in limited quantity, which often impedes the use of data-intensive machine learning approaches.

There are however situations cases where even if such data are unavailable, it is possible to artificially generate pairs by generating the input-output pairs by leveraging knowledge of the forward model (Eq. \eqref{eq:forward}), even if the latter is only approximately known. This involves generating a parameter/input vector $x$ sampled from the range of possible input values in the model or based on the known distribution of input data. We propose to compute the corresponding simulated observation $\hat{f}(x)$, i.e.  without noise. It is important to note that the forward model serves anyway as an approximation of the underlying physical phenomenon $f$ it represents, and the simulated observation can only be considered as a perturbed version of the unknown observation due to model imperfections.

Summarizing our discussion above, we propose to train a neural network $\psi$ by replacing Eq. \eqref{eq:trainpinnunsup} with
\begin{equation}\label{eq:trainpinnsup}
\min_{\theta \in \Theta} \frac{1}{n} \sum_{i=1}^n \mathcal{L}_{\lambda}(x_i,y_i,\theta) , {\rm where}
\end{equation}
\[
\mathcal{L}_{\lambda}(x,y,\theta) = \lambda\lVert y - \hat{f}(\psi(y,\theta)) \rVert^2 + (1-\lambda) \lVert \psi(y,\theta) - x \rVert^2 .
\]
Here $\lambda \in ]0,1[$ balances between the two terms: fidelity to the observation and reconstruction error. The determination of an optimal value for $\lambda$ can be a challenging task. In our study, we employed an empirical approach to estimate this value by performing cross-validation.

In the case of real observations, the value of $x$ is unknown, making it impossible to calculate the second term in the loss function (Eq. \eqref{eq:trainpinnsup}). Therefore, only the first term, which focuses on reconstruction fidelity, is utilized for such data.
 
The ratio between the number of real data, denoted as $N_o$ (where only the observation is known), and the number of simulated data, denoted as $N_s$, plays a significant role in the analysis. Consequently, the influence of this ratio have to be thoroughly examined through experimental studies.

\section{Experimental results}
\label{sec:experimental_validation}
% \todo{Qq'un des pbs inverses voit $n \ll m$, pensera clairement que le pb n'est pas difficile (plus d'équation que d'inconnues). Mais en fait, ici le pb est non-linéaire et on a des classes d'équivalence non-triviales. Il faudra donc insister sur ce message: on veut résoudre un pb inverse non-linéaire, où certes $n \ll m$, mais qui reste très mal-posé.}
% \rcom{}{We validate the proposed method on a source separation problem where the dimensions of $\mathcal{X}$ is relatively
% smaller compared to the dimensions of $\mathcal{Y}$ ($n \ll m$).
% In what follows, we present the problem in question. We then present the experimental settings and the results we got on this problem.}

% \todo{Autre point important, il faudra absolument ajouter un exemple issu de WatchTower, et motiver le travail dès le départ par cet exemple. Sinon, je doute que le papier ait sa place dans WHISPERS puisqu'il s'agit d'une conférence orienté télédétection.}

We validate the proposed method by applying it to an orbit restitution problem, where the dimensionality of $\mathcal{X}$ ($n = 6$) is relatively smaller than the dimensionality of $\mathcal{Y}$ ($m = 64^2$). This problem encompasses several intriguing aspects that make it particularly compelling. Firstly, the underlying physics and the involved forward operator exhibit nonlinearity, which is a primary focus in real-world research problems. Secondly, an orbit is defined by six orbital elements, while the received data exists in a significantly larger space, such as the image space in our case. Consequently, the forward operator maps from a smaller parameter space to a substantially larger image space. The third aspect of this problem pertains to the challenging nature of obtaining labelled data, as it requires integrating raw acquisition with non-trivial evaluations and determinations of orbit parameters.

It is important to note that despite initially appearing simple due to the presence of more equations than unknown variables, this problem poses additional difficulties. The involved physics operator is nonlinear, rendering the problem ill-posed with non-trivial equivalence classes. This complexity makes it more challenging than it may seem. In the subsequent sections, we present the details of the problem, including the experimental settings and the obtained results.

\subsection{Problem and dataset }

The objective is to derive an inverse operator for an orbit propagator using images obtained from a simulated sensor.

\begin{figure}[tb]
\begin{center}
\includegraphics[width=.4\textwidth]{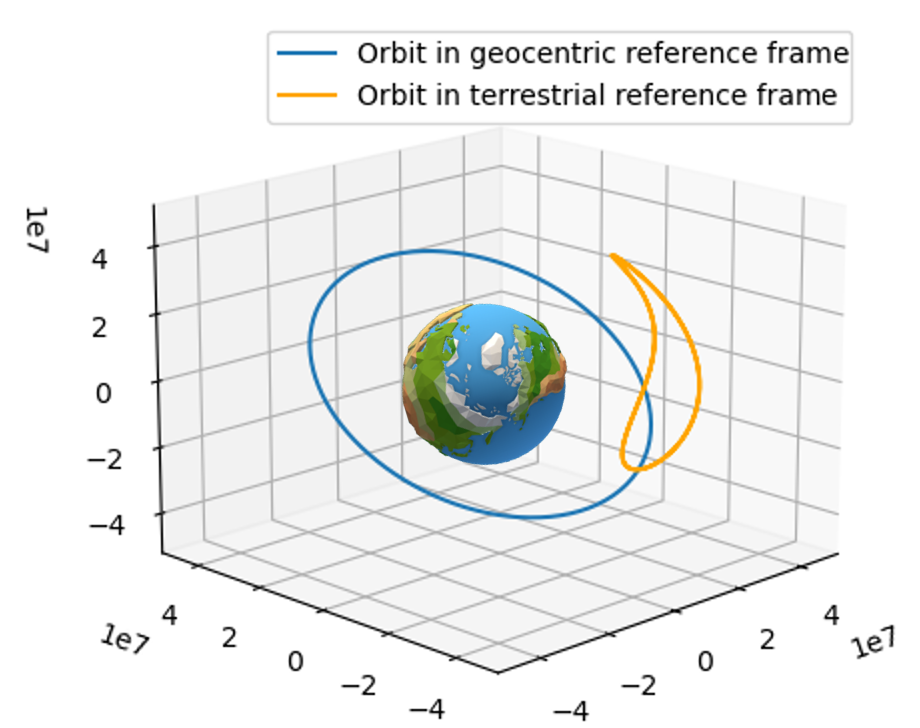}
\end{center}
\caption{Orbit Restitution: Visualization of an orbit in multiple reference frames. In this example, the depicted orbit has an inclination of 45 degrees, an eccentricity of 0.4, and an apogee of $42164\times 10^3$ meters. \label{fig:orbit_ref} } 
\end{figure}

\begin{figure}[tb]
\begin{center}
\includegraphics[width=4cm]{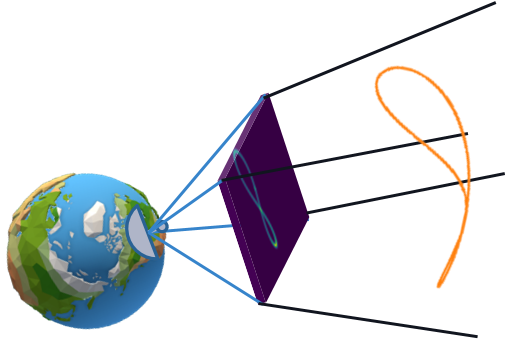}
\includegraphics[width=4cm]{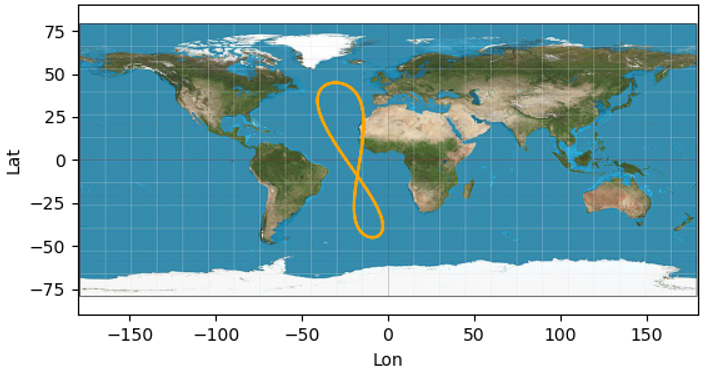}
\end{center}
\caption{Orbit restitution: Simulation of the orbit projection, as depicted in Figure \ref{fig:orbit_ref}. The left image exhibits the simulated sensor exhibiting the projected orbit, while the right image depicts the orbit projected onto the ground plane. \label{fig:orbit_proj_ex_fig}} 
\end{figure}

The forward operator performs the projection of orbits expressed in the Terrestrial Reference Frame (TRF) onto an image, similar to a ground track orbit projection. Consequently, the received data correspond to the projection of an orbit onto a sensor positioned on the Earth (see Figures \ref{fig:orbit_ref} and \ref{fig:orbit_proj_ex_fig}).

For this problem, we opt to represent each orbit using 3 Keplerian parameters: inclination, eccentricity, and periapsis. This choice leads to a wide range of images and results, while avoiding trivial equivalence classes in the parameter set.

In this problem, the forward operator as follows: 

\begin{equation}
\begin{array}{ccccc}
f &:&  [0,1[ \times  [0,2\pi[ \times  [0,2\pi[ & \to &  \mathcal{I}_{64\times64}(\mathbb{R})\\
 & &  e,i, \omega & \mapsto & .
\end{array}
\end{equation}

This operator simulates the entire system, encompassing the satellite position (achieved through an orbital propagator), the radiation pattern, and the projection to the final image (assumed to be $64\times64$ pixels). In this experiment, it is assumed that the approximated and actual physics models are identical, denoted as $\hat{f} = f$. To compute the satellite position while preserving the differentiability criterion of $\hat{f}$, an analytical Keplerian propagator is employed.

Finally, the neural network utilized for this problem consists of 5 dense layers, with each layer containing 784 neurons and employing a ReLU activation function. To assess the performance of the neural network, two previously defined loss functions are employed, as illustrated in Figure \ref{fig:proposed_models}. Specifically, the Mean Squared Error (MSE) is utilized for both the reconstruction error and the parameter error evaluation.

\subsection{Results and analysis}
\label{sec:results_and_analysis}

\begin{table}[tb]
    \begin{center}
        \renewcommand*{\minval}{0.4121}
        \renewcommand*{\maxval}{2.4541}
    \resizebox{\columnwidth}{!}{\begin{tabular}{ |c|c||c|c|c|c|c|} 
        \cline{3-7}
        \multicolumn{2}{c|}{\multirow{2}{*}{ArgPer($\times10^{-5}$)}}& \multicolumn{5}{c|}{Number of real observations ($N_o$)}\\ \cline{3-7}
        \multicolumn{2}{c|}{}&0&1000&10000&20000&40000\\ \cline{3-7} \noalign{\vskip\doublerulesep
               \vskip-\arrayrulewidth} \hline
       \multirow{5}{*}{\rotatebox[origin=c]{90}{$N_s$}} &0&-&\gradient{2.4541}&\gradient{1.2970}&\gradient{0.9947}&\gradient{0.9081}\\ \cline{2-7}
          & 1000& \gradient{2.0546} &\gradient{2.3412}&\gradient{1.1546}&\gradient{0.9731}&\gradient{1.1742}  \\ \cline{2-7}
          & 10000& \gradient{1.1470}&\gradient{2.1942}&\gradient{0.8591}&\gradient{0.7854}&\gradient{1.0644}\\ \cline{2-7}
          & 20000&\gradient{0.8401} &\gradient{0.7721}&\gradient{0.6924}&\gradient{0.6571}&\gradient{0.6431} \\ \cline{2-7}
          & 40000& \gradient{0.4541}&\gradient{0.4412}&\gradient{0.4201}&\gradient{0.4121}&\gradient{0.4212}\\ \hline
      \end{tabular}}
        \renewcommand*{\minval}{0.2861}
        \renewcommand*{\maxval}{1.6556}
    \resizebox{\columnwidth}{!}{
        \begin{tabular}{ |c|c||c|c|c|c|c|} 
            \cline{3-7}
            \multicolumn{2}{c|}{\multirow{2}{*}{\small Eccentricity($\times10^{-5}$)}}& \multicolumn{5}{c|}{Number of real observations ($N_o$)}\\ \cline{3-7}
            \multicolumn{1}{c}{}& \multicolumn{1}{c|}{}&0&1000&10000&20000&40000\\ \cline{3-7} \noalign{\vskip\doublerulesep
                    \vskip-\arrayrulewidth} \hline
            \multirow{5}{*}{\rotatebox[origin=c]{90}{$N_s$}} &0&- &\gradient{1.5587}& \gradient{1.4424} & \gradient{1.1232}& \gradient{0.8638}\\ \cline{2-7}
                & 1000& \gradient{1.3545} &\gradient{1.5225}&\gradient{0.7542}&\gradient{1.1036}&\gradient{0.8452} \\ \cline{2-7}
                & 10000& \gradient{1.6556}& \gradient{1.5245}& \gradient{0.7054}&\gradient{1.1023}& \gradient{0.7214}\\ \cline{2-7}
                & 20000& \gradient{1.4684}&\gradient{1.0845}&\gradient{1.2781}&\gradient{0.6306}&\gradient{0.6251} \\ \cline{2-7}
                & 40000& \gradient{0.3895} &\gradient{0.3856}& \gradient{0.3776}&\gradient{0.3435}&{\gradient{0.2861}}\\ \hline
        \end{tabular}}
              
        \renewcommand*{\minval}{0.0001}
        \renewcommand*{\maxval}{0.0134}
\resizebox{\columnwidth}{!}{\begin{tabular}{ |c|c||c|c|c|c|c|} 
    \cline{3-7}
   \multicolumn{2}{c|}{\multirow{2}{*}{Inclination}}& \multicolumn{5}{c|}{Number of real observations ($N_o$)}\\ \cline{3-7}
    \multicolumn{2}{c|}{}&0&1000&10000&20000&40000\\ \cline{3-7} \noalign{\vskip\doublerulesep
           \vskip-\arrayrulewidth} \hline
   \multirow{5}{*}{\rotatebox[origin=c]{90}{$N_s$}} &0&-&\gradient{0.0132}&\gradient{0.0021}&\gradient{0.0009}&\gradient{0.0004}\\ \cline{2-7}
      & 1000&  \gradient{0.0134}&\gradient{0.0114}&\gradient{0.0017}&\gradient{0.0012}&\gradient{0.0003}  \\ \cline{2-7}
      & 10000& \gradient{0.0022}&\gradient{0.0028}&\gradient{0.0024}&\gradient{0.0008}&\gradient{0.0007}\\ \cline{2-7}
      & 20000& \gradient{0.0014}&\gradient{0.0007}&\gradient{0.0007}&\gradient{0.0007}&\gradient{0.0005} \\ \cline{2-7}
      & 40000& \gradient{0.0002}&\gradient{0.0002}&\gradient{0.0002}&\gradient{0.0001}&\gradient{0.0001}\\ \hline
  \end{tabular}}

    \renewcommand*{\minval}{0.0251}
        \renewcommand*{\maxval}{1.4581}
            \resizebox{\columnwidth}{!}{\begin{tabular}{ |c|c||c|c|c|c|c|} 

  \cline{3-7}
 \multicolumn{2}{c|}{\multirow{2}{*}{\small Reconstruction}}& \multicolumn{5}{c|}{Number of real observations ($N_o$)}\\ \cline{3-7}
 \multicolumn{2}{c|}{}&0&1000&10000&20000&40000\\ \cline{3-7} \noalign{\vskip\doublerulesep
         \vskip-\arrayrulewidth} \hline
 \multirow{5}{*}{\rotatebox[origin=c]{90}{$N_s$}} &0&-&\gradient{1.3243}&\gradient{1.1667}&\gradient{0.8742}&\gradient{0.3023}\\ \cline{2-7}
    & 1000&  \gradient{1.4581}&\gradient{0.7064}&\gradient{0.1314}&\gradient{0.0989}&\gradient{0.0852}  \\ \cline{2-7}
    & 10000& \gradient{0.2467}&\gradient{0.1510}&\gradient{0.1394}&\gradient{0.0920}&\gradient{0.0955}\\ \cline{2-7}
    & 20000& \gradient{0.0966}&\gradient{0.0831}&\gradient{0.0968}&\gradient{0.0750}&\gradient{0.0701} \\ \cline{2-7}
    & 40000& \gradient{0.0304}&\gradient{0.0305}&\gradient{0.0298}&\gradient{0.0271}&{\gradient{0.0251}}\\ \hline
\end{tabular}}
    \end{center}
    \caption{
Error evaluation across various parameters, using a test set comprising solely of observations.  \label{tab:orbite_results}}
    \end{table}

 % As one can see in the Table \ref{tab:orbite_results}, the SimPINNs method can handle and extract the information given by the parameter loss and the image loss converge to a better minimum. And achieves better performance in terms of parameter reconstruction than the PINNs method (represented by the dataset composed of 0 generated data and 40k real data). In fact, the best results are obtained with the largest number of real and generated data with an average factor of $\times 2$ for the parameter error compared to the unsupervised PINNs method (resp. a factor of $ \times10$ for the reconstruction error). 

 As shown in Table \ref{tab:orbite_results}, the SimPINNs method effectively utilizes and leverages information from both real observations and generated data. It outperforms the PINNs method in terms of parameter reconstruction; PINNs is the row with $N_s=0$. The most favorable results are achieved when employing the largest number of real and generated data points ($N_o = N_s = 40k$), with an average improvement of a factor of 2 for the parameter error compared to the unsupervised PINNs method. Additionally, the SimPINNs method demonstrates a significant advantage, reducing the reconstruction error by a factor of 10.
 
The impact of incorporating orbital physics in the forward operator becomes evident in the significant benefit it provides for image reconstruction in this particular use case. Due to the influence of orbital dynamics, even small changes in the parameters can have a substantial impact on the satellite's orbit and drastically alter the resulting projection on the observed image. As a result, two images with nearly identical parameters can exhibit significant differences. In this context, the reconstruction loss plays a crucial role in assisting the network in handling these high-gradient values that may not be adequately captured by the supervised loss alone.

By combining both the generated and real data, the model can benefit from regularization through $\mathcal{L}_\lambda$ applied to the real dataset. This approach allows the model to leverage the strengths of both data types and converge towards an improved solution with reduced parameter error.

\begin{figure}[htp!]
\begin{center}
\includegraphics[scale=0.50]{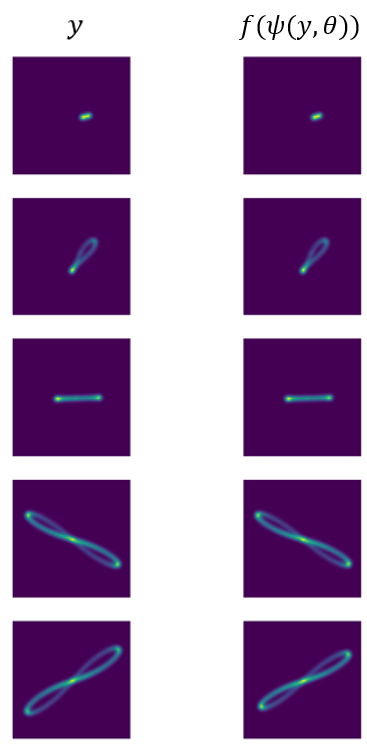}
\end{center}
\caption{Reconstruction Results using the SimPINNs Method. \label{fig:orbit_proj_ex}}
\end{figure}

Figure \ref{fig:orbit_proj_ex} presents a selection of projected orbits along with their corresponding reconstructions.

\section{Conclusions}
\label{sec:conclusion}

This article explores an NN approach to solve non-linear inverse problems, focusing on the promising SimPINNs method. SimPINNs leverages the physics model to generate labeled training data and effectively regularize the neural network. The study demonstrates that simulation-aided training provides more information compared to conventional PINNs or vanilla neural network training. By utilizing the approximated physics operator, the model achieves improved learning and generalization over non-labeled datasets. SimPINNs shows potential for addressing challenging inverse problems with limited labeled data, offering insights into training neural networks with physics-based knowledge. 

\bigskip
\noindent {\sc \large \bf Acknowledgements}\\[1em]
The research presented in this paper is, in part, funded by the French National Research Agency (ANR) through the grant ANR-19-CHIA-0017-01-DEEP-VISION.

% However, the case in which the approximated and actual physics operators do not match perfectly has not been experimentally investigated, and it would be interesting to see how the method performs in such a scenario.

% This is even more accurate when the true and forward physics model are slightly different. When the true and forward physics models differ marginally, this is even more accurate. In this instance, simulation-assisted training provides the best of both worlds: regularization of the solution space from labelled data and generalization over real-world data.

% \vfill
% \pagebreak

% \section{REFERENCES}
% \label{sec:ref}

% List and number all bibliographical references at the end of the paper.  The references can be numbered in alphabetic order or in order of appearance in the document.  When referring to them in the text, type the corresponding reference number in square brackets as shown at the end of this sentence \cite{C2}.

% References should be produced using the bibtex program from suitable
% BiBTeX files (here: strings, refs, manuals). The IEEEbib.bst bibliography
% style file from IEEE produces unsorted bibliography list.
% -------------------------------------------------------------------------
\bibliographystyle{IEEEbib}
\bibliography{refs}

\end{document}